%% file: main.tex
\documentclass{article}

    \PassOptionsToPackage{numbers, compress}{natbib}

\usepackage{relsize}
\usepackage{multicol}
\usepackage{multirow}

    \usepackage[preprint]{neurips_2023}

\usepackage[utf8]{inputenc} 
\usepackage[T1]{fontenc}    
\usepackage{hyperref}       
\usepackage{url}            
\usepackage{booktabs}       
\usepackage{amsfonts}       
\usepackage{nicefrac}       
\usepackage{microtype}      
\usepackage{xcolor}         
\usepackage{graphicx}
\usepackage{subfigure}
\usepackage{bm}
\usepackage{amsmath}

\title{Atomic and Subgraph-aware Bilateral Aggregation for Molecular Representation Learning}

\author{%
  Jiahao Chen\textsuperscript{\rm 1 2},\quad Yurou Liu\textsuperscript{\rm 1 2},\quad Jiangmeng Li\textsuperscript{\rm 3 4},\quad Bing Su\textsuperscript{\rm 1 2}\thanks{Corresponding author},\quad Jirong Wen\textsuperscript{\rm 1 2} \\
  \textsuperscript{\rm 1}Gaoling School of Artificial Intelligence, Renmin University of China \\
  \textsuperscript{\rm 2}Beijing Key Laboratory of Big Data Management and Analysis Methods\\
  \textsuperscript{\rm 3}Institute of Software Chinese Academy of Sciences\\
  \textsuperscript{\rm 4}University of Chinese Academy of Sciences\\
  \texttt{\{nicelemon666, yurouliu99, subingats\}@gmail.com}\\
  \texttt{jiangmeng2019@iscas.ac.cn, jrwen@ruc.edu.cn} 
}

\begin{document}

\maketitle

\begin{abstract}
Molecular representation learning is a crucial task in predicting molecular properties.  Molecules are often modeled as graphs where atoms and chemical bonds are represented as nodes and edges, respectively, and Graph Neural Networks (GNNs) have been commonly utilized to predict atom-related properties, such as reactivity and solubility. However, functional groups (subgraphs) are closely related to some chemical properties of molecules, such as efficacy, and metabolic properties, which cannot be solely determined by individual atoms. 
In this paper, we introduce a new model for molecular representation learning called the Atomic and Subgraph-aware Bilateral Aggregation (ASBA), which addresses the limitations of previous atom-wise and subgraph-wise models by incorporating both types of information. ASBA consists of two branches, one for atom-wise information and the other for subgraph-wise information. Considering existing atom-wise GNNs cannot properly extract invariant subgraph features, we propose a decomposition-polymerization GNN architecture for the subgraph-wise branch. Furthermore, we propose cooperative node-level and graph-level self-supervised learning strategies for ASBA to improve its generalization. Our method offers a more comprehensive way to learn representations for  molecular property prediction and has broad potential in drug and material discovery applications. Extensive experiments have demonstrated the effectiveness of our method.
\end{abstract}

\input{0.introduction}
\input{1.related_work}
\input{2.method}
\input{3.experiment}
\input{4.conclusion}

\bibliographystyle{plain}  
\bibliography{references}








\end{document}

%% file: 0.introduction.tex
\section{Introduction}
Molecular representation learning plays a fundamental role in predicting molecular properties, which has broad applications in drug and material discovery~\cite{feinberg2018potentialnet}. Previous methods typically model molecules as graphs, where atoms and chemical bonds are modeled as nodes and edges, respectively. Therefore, Graph Neural Networks (GNNs)~\cite{hamilton2017inductive} have been widely applied to predict specific properties associated with atoms, such as solubility and reactivity~\cite{zhang2021motif, hu2019strategies, yang2022learning}. For simplicity, we refer to these models as atom-wise models, as the final molecular representation is the average of each atom.

\begin{figure}
    \center
    \subfigure[ASBA]{\includegraphics[width=0.48\textwidth]{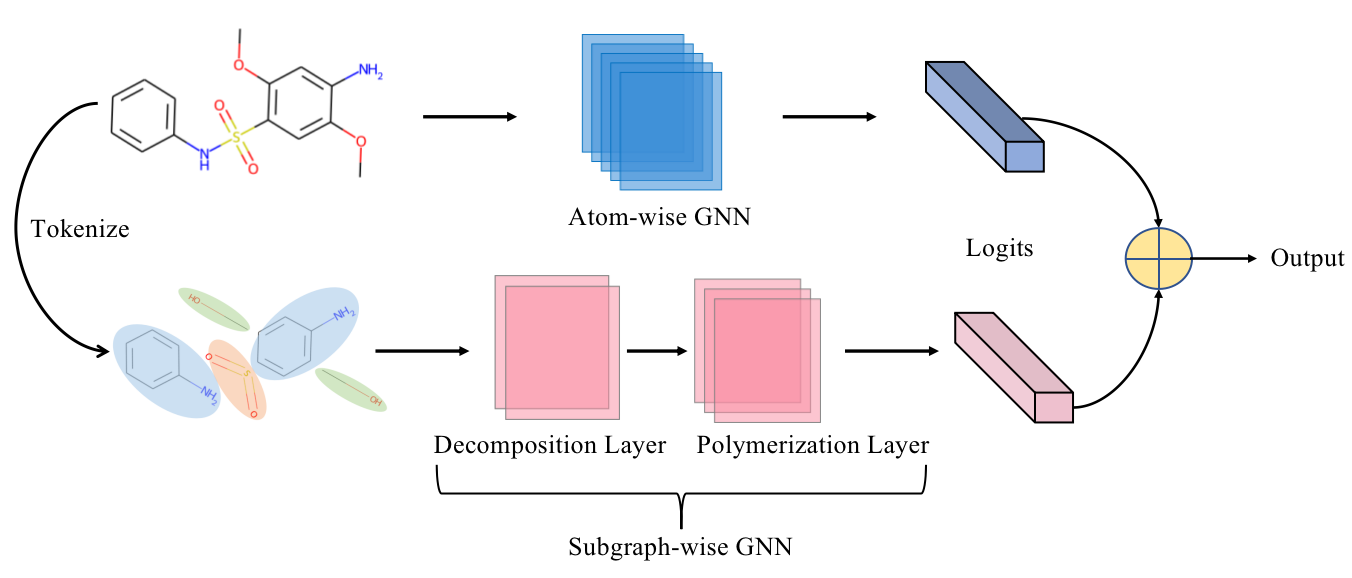}}
    \subfigure[BBBP]{\includegraphics[width=0.25\textwidth]{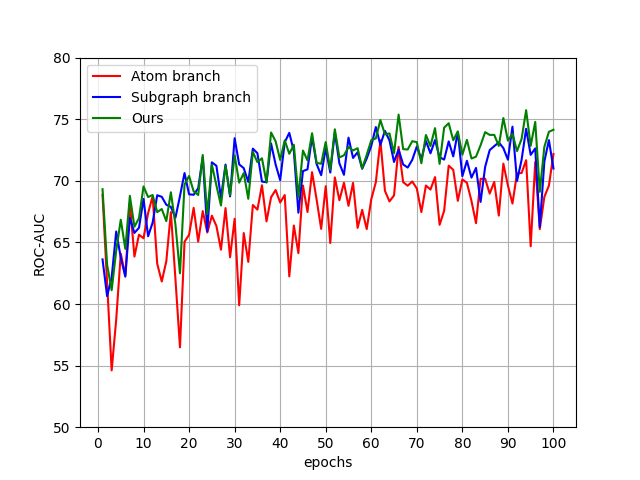}}
    \subfigure[ToxCast]{\includegraphics[width=0.25\textwidth]{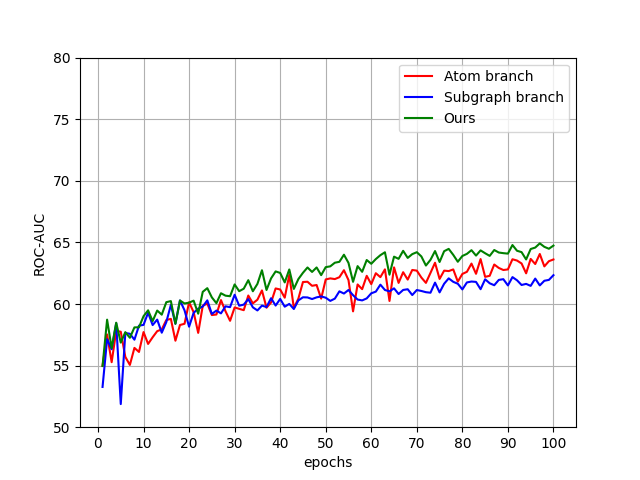}}
    \caption{(a) Overview of our ASBA. (b) Testing curves on the BBBP dataset. (c) Testing curves on the ToxCast dataset.}
    \label{fig1}
\end{figure}

However, not all molecular properties are determined by individual atoms, and some chemical properties are closely related to functional groups (subgraphs)~\cite{yang2022learning, kong2022molecule}. For example, a molecule's efficacy, and metabolic properties are often determined by the functional groups within it. Therefore, many methods propose subgraph-wise models~\cite{jiang2023pharmacophoric, yang2022learning}, where the final molecular representation is the average of each subgraph. The method proposed in~\cite{yang2022learning} decomposes a given molecule into pieces of subgraphs and learns their representations independently, and then aggregates them to form the final molecular representation. By paying more attention to functional groups, this method ignores the influence of individual atoms\footnote{The representation of each subgraph merges the atom-wise representation by the attention mechanism, but it does not model atom-representation relationships explicitly.}, which is harmful to predicting properties related to atoms. More commonly,  different properties are sensitive to atoms or functional groups but often determined by both of them at the same time. To verify our point, we do experiments on BBBP and ToxCast dataset and visualize the performance on the test set in Fig.~\ref{fig1} (b) and (c), respectively. It is shown that BBBP is more sensitive to the subgraph-wise branch while Toxcast is the atom-wise branch. 

Therefore, both atom-wise and subgraph-wise models have inadequacies and cannot accurately predict current molecular properties independently. To address this dilemma, we propose our Atomic and Subgraph-aware Bilateral Aggregation (ASBA). As shown in Fig.~\ref{fig1} (a), ASBA has two branches, one modeling atom-wise information and the other subgraph-wise information. We follow previous work in constructing the atom-wise branch, which aggregates the representation of each atom and employs a linear classifier to predict properties. For the subgraph-wise branch, we propose a Decomposition-Polymerization GNN architecture where connections among subgraphs are broken in the lower decomposition GNN layers and each subgraph is viewed as a single node in the higher polymerization layers. The decomposition layers serve as a separate GNN module that embeds each subgraph into an embedding, and the relations among subgraphs are modeled by polymerization layers. In this way, the second branch takes subgraphs as basic token units and aggregates them as the final representation. Similar to the atom-wise branch, an independent classifier is added to predict the property. Finally, we incorporate the outputs of the two branches as the prediction score.

In addition, we propose a corresponding self-supervised learning method to jointly pre-train the two branches of our ASBA. Existing self-supervised molecular learning methods mainly design atom-wise perturbation invariance or reconstruction tasks, e.g. predicting randomly masked atoms, but they cannot fully capture subgraph-wise information and the relations among substructures. To this end, we propose a Masked Subgraph-Token Modeling (MSTM) strategy for the subgraph-wise branch. MSTM first tokenizes a given molecule into pieces and forms a subgraph-token dictionary. Compared with atom tokens, such subgraphs correspond to different functional groups, thus their semantics are more stable and consistent. MSTM decomposes each molecule into subgraph tokens, masks a portion of them, and learns the molecular representation by taking the prediction of masked token indexes in the dictionary as the self-supervised task. For the atom-wise branch, we simply employ the masked atom prediction task. Although the atom-wise branch and the subgraph-wise branch aim to extract molecular features from different levels, the global graph representations for the same molecule should be consistent. To build the synergistic interaction between the two branches for joint pre-training, we perform contrastive learning to maximize the consistency of representations between different branches. Experimental results show the effectiveness of our method.



Our contributions can be summarized as:
\begin{enumerate}
\item We propose a bilateral aggregation model to encode the characteristics of both atoms and subgraphs with two branches. For the subgraph branch, we propose a novel decomposition-polymerization architecture to embed each subgraph token independently with decomposition layers and polymerize subgraph tokens into the final representation with polymerization layers. In this way, subgraph-level embeddings remain invariant in different molecules, and structural knowledge is explicitly incorporated.
\item We propose a cooperative node-level and graph-level self-supervised learning method to jointly train the two branches of our bilateral model. For the subgraph branch, we propose MSTM, a novel self-supervised molecular learning strategy, which uses the auto-discovered subgraphs as tokens and predicts the dictionary indexes of masked tokens. The subgraph tokens are more stable in function and have more consistent semantics. In this way, masked subgraph modeling can be performed in a principled manner. At the global graph level, we perform a contrastive learning strategy that imposes the interaction of the two branches with the consistency constraint.
\item We provide extensive empirical evaluations to show that the learned representation by our bilateral model and our self-supervised learning method has a stronger generalization ability in various functional group-related molecular property prediction tasks.
\end{enumerate}

%% file: 1.related_work.tex
\section{Related work}
\paragraph{Molecular Property Prediction}
The prediction of molecular properties is an important research topic in the fields of chemistry, materials science, pharmacy, biology, physics, etc~\cite{wang2011application}. Since it is time-consuming and labor-intensive to measure properties via traditional wet experiments, many recent works focus on designing end-to-end machine learning methods to directly predict properties. These works can be divided into two categories: SMILES string-based methods~\cite{butler2018machine,dong2018admetlab} and graph-based methods~\cite{gilmer2017neural,yang2019analyzing,lu2019molecular,gasteiger2020directional}. Compared with SMILES strings, it is more natural to represent molecules as graphs and model them with Graph neural networks (GNNs). However, the training of GNNs requires a large amount of labeled molecule data and supervised-trained GNNs usually show limited generalization ability for newly synthesized molecules and new properties. In order to tackle these issues, self-supervised representation pre-training techniques are explored~\cite{rong2020self,li2021effective,stark20223d} in molecular property prediction.

\paragraph{Self-supervised Learning of Graphs}
Based on how self-supervised tasks are constructed, previous works can be classified into two categories, contrastive models and predictive models. Contrastive models~\cite{hu2019strategies,zhang2020motif,sun2020infograph,you2021graph,sun2021mocl,subramonian2021motif,xia2022simgrace,li2022let,suresh2021adversarial} generate different views for each graph via data augmentation and learn representations by contrasting the similarities between views of the same graph and different graphs. Predictive models \cite{hu2020gpt,rong2020self,hou2022graphmae} generally mask a part of the graph and predict the masked parts. Most existing methods focus on learning node-level or graph-level representations, with some work involving subgraph-level which utilizes the rich semantic information contained in the subgraphs or motifs. For instance, in~\cite{zhang2021motif}, the topology information of motifs is considered. In~\cite{wu2023molformer}, a Transformer architecture is proposed to incorporate motifs and construct 3D heterogeneous molecular graphs for representation learning. Different from these works, we not only propose a bilateral aggregation model with a novel subgraph-aware GNN branch but also propose a joint node-wise and graph-wise self-supervised training strategy so that the learned representation can capture both atom-wise and subgraph-wise information.

%% file: 2.method.tex
\section{Methodology}
\subsection{Problem formulation}
We represent a molecule as a graph $G=<V, E>$ with node attribute vectors $\bm{x}_v$ for $v \in V$ and edge attribute vectors $\bm{e}_{uv}$ for $(u,v) \in E$, where $V$ and $E$ are the sets of atoms and bonds, respectively. We consider a multilabel classification problem with instance $G$ and label $\bm{y} = \{y_l\}_{l=1}^L$, where $y_l \in \{0, 1\}$ denotes whether this property is present in $G$. Given a set of training samples $\mathcal{D}_{train}=\{(G_i, \bm{y}_i)\}_{i=1}^{N_1}$, our target is to learn a mapping $f: G \rightarrow \mathbb{R}^L$ that can well generalize to the test set. We also have a set of unlabelled support set $\mathcal{D}_{support}=\{(G_i)\}_{i=1}^{N_2}$, where $N_2 >> N_1$, and apply our self-supervised learning method to get better initial representation. 

\subsection{Atom-wise branch}
Previous works extract the representation of a molecule by aggregating the embeddings of all atoms with GNNs. Similarly, our atom-wise branch applies a single GNN model with $K$ layers to map each molecule graph into an embedding. Specifically, for $G=<V, E>$, the input embedding $\bm{h}_{v}^0$ of the node $v \in V$ is initialized by $\bm{x}_v$, the input embedding at the $k$-th layer $\bm{e}_{uv}^k$ of the edge $(u, v) \in E$ is initialized by $\bm{e}_{uv}$, and the $K$ GNN layers iteratively update $\bm{h}_{v}$ by polymerizing the embeddings of neighboring nodes and edges of $\hat{v}$. In the $k$-th layer, $\bm{h}_{v}^{(k)}$ is updated as follows:
\begin{equation}
\begin{aligned}
\bm{h}_{v}^{(k)} = {\rm COMBINE}^{(k)}(\bm{h}_{v}^{(k-1)},{\rm AGGREGATE}^{(k)}(\{(\bm{h}_{v}^{(k-1)},\bm{h}_{u}^{(k-1)},\bm{e}_{uv}^k):u \in \mathcal{N}(v) \}))
\end{aligned}
\end{equation}
where $\bm{h}_{v}^{(k)}$ denotes the embedding of node $v$ at the $k$-th layer, and $\mathcal{N}(v)$ represents the neighborhood set of node $v$. After $K$ iterations of aggregation, $\bm{h}_{v}^{(K)}$ captures the structural information within its $K$-hop network neighborhoods. The embedding $\bm{z}_{A}$ of the graph $G$ is the average of each node.
\begin{equation}
\bm{z}_{A} = {\rm MEAN}(\{\bm{h}_{v}^{(K)}|v \in V\})
\label{eq1}
\end{equation}
Then we add a linear classifier $\bm{z}_A \rightarrow \mathbb{R}^L$. Formally, atom-wise architecture can be described as learning a mapping $f: G \rightarrow \mathbb{R}^L$. The loss function of the atom-wise architecture is:
\begin{equation}
    \mathcal{L}_{atom} =  \frac{1}{N_1}\sum_{(G, \bm{y}) \in \mathcal{D}_{train}}  \ell\Big(f(G), \bm{y}\Big)
\end{equation}

\subsection{Subgraph-wise branch}
Atoms are influenced by their surrounding contexts and the semantics of a single atom can change significantly in different environments. Functional groups, which are connected subgraphs composed of coordinated atoms, determine most molecular properties. Our proposed hierarchical Decomposition-Polymerization architecture decouples the representation learning into the subgraph embedding phase, where each molecule is decomposed into subgraphs and an embedding vector is extracted from each subgraph, and the subgraph polymerization phase, where subgraphs are modeled as nodes and their embeddings are updated by polymerizing information from neighboring subgraphs. Finally, the final representation is obtained by combining all subgraph-wise embeddings.

\paragraph{Subgraph vocabulary construction}
Functional groups correspond to special subgraphs, however, pre-defined subgraph vocabularies of hand-crafted functional groups may be incomplete, i.e., not all molecules can be decomposed into disjoint 
subgraphs in the vocabulary. There exist many decomposition algorithms such as the principle subgraph extraction strategy~\cite{kong2022molecule} and breaking retrosynthetically interesting chemical substructures (BRICS)~\cite{degen2008art}. Generally,  we denote a subgraph of the molecule $G$ by $S=<\hat{V}, \hat{E}> \in G$, where $\hat{V}$ is a subset of $V$ and $\hat{E}$ is the subset of $E$ corresponding to $\hat{V}$. The target of principle subgraph extraction is to constitute a vocabulary of subgraphs $\mathbb{V}=\{S_{(1)}, S_{(2)}, \cdots, S_{(M)}\}$ that represents the meaningful patterns within molecules, where each unique pattern is associated with an index. Details are discussed in Appendix.

\paragraph{Subgraph embedding}
In this phase, we only focus on learning the embedding of each subgraph by modeling the intra-subgraph interactions. For a molecule $G=<V, E>$, we decompose it into a set of non-overlapped subgraphs $\{S_{\pi 1}, S_{\pi 2}, \cdots, S_{\pi T}\}$, where $T$ is the number of decomposed subgraphs and $\pi t$ is the corresponding index of the $t^{th}$ decomposed subgraph in the constructed vocabulary $\mathbb{V}$. For each subgraph $S_{\pi t}=<\hat{V}_{\pi t}, \hat{E}_{\pi t}>$, we have $\hat{V}_{\pi t} \subset V$ and $\hat{E}_{\pi t} \subset E$. For each edge $(u, v)$ in $E$, we add it into the inter-subgraph edge set $\mathcal{E}$ if it satisfies that nodes $u$ and $v$ are in different subgraphs. Therefore, we have $V = \cup \hat{V}_{\pi t}$ and $E = \cup \hat{E}_{\pi t} \cup \mathcal{E}$. 

We apply a single GNN model with $K_1$ layers to map each decomposed subgraph into an embedding. GNN depends on the graph connectivity as well as node and edge features to learn an embedding for each node $v$. We discard the inter-subgraph edge set $\mathcal{E}$, any two subgraphs are disconnected and the information will be detached among subgraphs. This is equivalent to feeding each subgraph $S_{\pi t}$ into the GNN model individually. 

By feeding the molecular graph after discarding all inter-subgraph edges into the GNN model, the embeddings of all atoms in the $T$ decomposed subgraphs are updated in parallel and the embeddings of all subgraphs can be obtained by adaptive pooling. Compared with previous strategies~\cite{hu2019strategies, zhang2021motif} that directly obtain molecular representations from the context-dependent atom-wise embeddings with all edges, our strategy first extracts subgraph-level embeddings. When a subgraph appears in different molecules, both its atom-wise embeddings and the subgraph embedding remain the same. 

\paragraph{Subgraph-wise polymerization} 
In the previous subgraph embedding phase, we view each atom in the subgraph as a node and extract the embedding of each subgraph. In the subgraph-wise polymerization phase, we polymerize the embeddings of neighboring subgraphs for acquiring representations of subgraphs and the final representation of the molecule $G$. Differently, we view each subgraph as a node and connect them by the set of inter-subgraph edges $\mathcal{E}$. Two subgraphs $S_{\pi t}$ and $S_{\pi l}$ are connected if there exists at least one edge $(\hat{u}, \hat{v}) \in \mathcal{E}$ where $\hat{u} \in \hat{V}_{\pi t}$ and $\hat{v} \in \hat{V}_{\pi l}$. In this way, we construct another graph whose nodes are subgraphs and employ another GNN model with $K_2$ layers to update the representation of each subgraph and extract the final representation $\bm{z}_S$. At the $k'$-th layer, the embedding $\bm{h}_{\pi t}$ for the $t$-th subgraph is updated as follows:
\begin{equation}
\begin{aligned}
\bm{h}_{\pi t}^{(k')} = {\rm COMBINE}^{(k')}(\bm{h}_{\pi t}^{(k'-1)},{\rm AGGREGATE}^{(k')}(\{\bm{h}_{\pi t}^{(k'-1)}, \bm{h}_{\pi l}^{(k'-1)},e_{\hat{u}\hat{v}}^{k'}): (\hat{u}, \hat{v}) \in \mathcal{E} \\ {\rm AND} \quad \hat{u} \in \hat{V}_{\pi t} \quad {\rm AND} \quad \hat{v} \in \hat{V}_{\pi l}\})
\end{aligned}
\end{equation}
As shown in Eq.\ref{eq2}, representation $\bm{z}_S$ has aggregated all information from different subgraphs, where
$\bm{h}_{\pi t}^{(K_2)}$ denotes the subgraph feature which is fed forward after $K_2$ iterations.
\begin{equation}
    \bm{z}_S = {\rm MEAN} (\{\bm{h}_{\pi t}^{(K_2)}|t \in \{1,2, \cdots, k\}\})
    \label{eq2}
\end{equation}
The semantics of subgraphs corresponding to functional groups are relatively more stable in different molecular structures. Our polymerization takes such subgraphs as basic units to model the structural interactions and geometric relationships between them. Similarly, we add a linear classifier $\bm{z}_S \rightarrow \mathbb{R}^L$. Formally, subgraph-wise architecture can be described as learning a mapping $g: G \rightarrow \mathbb{R}^L$. The final loss function is:
\begin{equation}
    \mathcal{L}_{subgraph} =  \frac{1}{N_1}\sum_{(G, \bm{y}) \in \mathcal{D}_{train}}  \ell\Big(g(G), \bm{y}\Big)
\end{equation}

\subsection{Atomic and Subgraph-aware Bilateral Aggregation (ASBA)}
In this section, we propose our ASBA architecture. In the training phase, atom-wise and subgraph-wise branches are trained independently. In the testing phase, since there exists some complementary information between atoms and subgraphs, we aggregate the output of two branches. Formally, the final output is $(f+g)(G) = \Bigl(f(G) + g(G)\Bigr) / 2$. 

We also give a comprehensive analysis in Sec.~\ref{ana} to explain the superiority of our strategy.

\begin{figure*}[t]
\center
\includegraphics[width=1.\columnwidth]{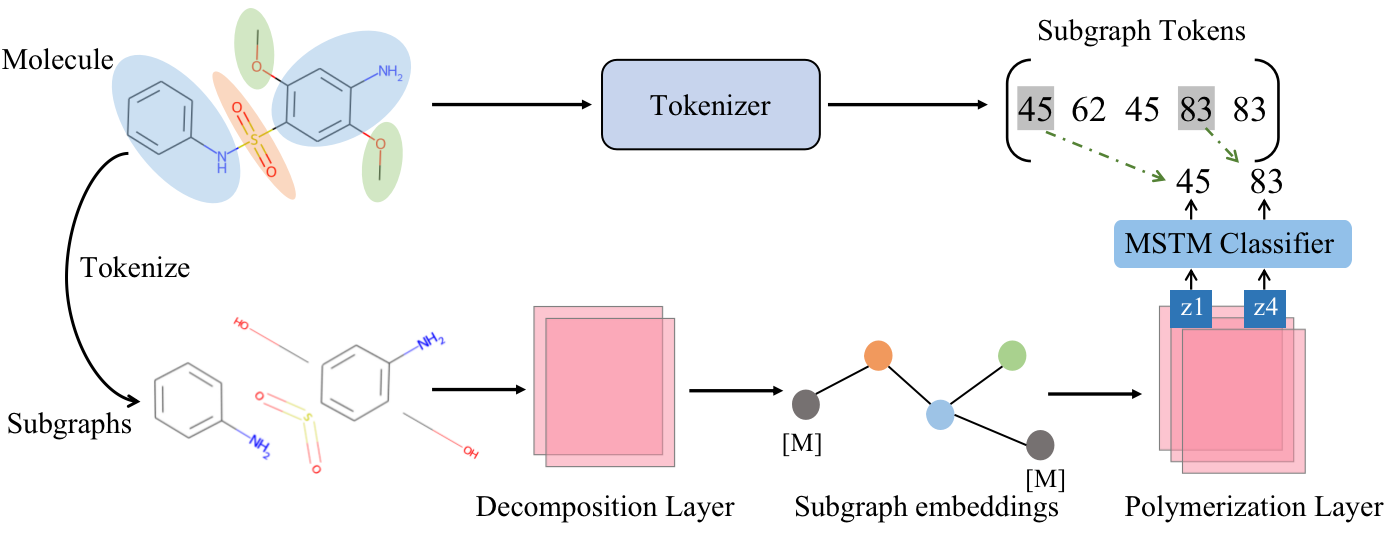}
\caption{Overview of the MSTM pre-training. We first construct a subgraph vocabulary by automatically mining principle subgraphs from data and each graph in the vocabulary is tokenized. During pre-training, each molecule is decomposed into subgraphs and then fed into the decomposition module, which maps each subgraph into an embedding. Some subgraphs are randomly masked and replaced with a special embedding. These subgraph embeddings are modeled as nodes and further fed into the polymerization module to extract subgraph-level representations. The pre-training task is to predict the masked tokens from the corrupted subgraph-level representations with a linear predictor.}
\label{image2}
\end{figure*}

\subsection{Self-supervised learning}
\paragraph{Node-level self-supervised learning}
Many recent works show that self-supervised learning can learn generalizable representations from large unlabelled molecules. Since the atom-wise branch and subgraph-wise branch are decoupled, we can easily apply the existing atom-wise self-supervised learning method to the atom-wise branch of ASBA such as attrMasking~\cite{hu2019strategies}.

For the subgraph-wise branch, we propose the Masked Subgraph-Token Modeling (MSTM) strategy, which randomly masks some percentage of subgraphs and then predicts the corresponding subgraph tokens. As shown in Fig.\ref{image2}, a training molecule $G$ is decomposed into $T$ subgraphs $\{S_{\pi 1}, S_{\pi 2}, \cdots, S_{\pi T}\}$. The subgraphs are tokenized to tokens $\{{\pi 1}, {\pi 2}, \cdots {\pi T}\}$, respectively, where $\pi t$ is the index of the $t$-th subgraph in the vocabulary $\mathbb{V}$. Similar to BEiT~\cite{bao2021beit}, we randomly mask a number of $M$ subgraphs and replace them with a learnable embedding. Therefore, we construct a corrupted graph $\tilde{G}$ and feed it into our hierarchical decomposition-polymerization GNN architecture to acquire polymerized representations of all subgraphs. For each masked subgraph $\tilde{S}_{\pi t}$, we bring an MSTM classifier $p(\cdot|h_{\pi t})$ with weight $\bm{W}_p$ and bias $\bm{b}_p$ to predict the ground truth token $\pi t$. Formally, the pre-training objective of MSTM is to minimize the negative log-likelihood of the correct tokens given the corrupted graphs.
\begin{equation}
    \mathcal{L}_{MSTM} = \frac{1}{N_2}
    \sum_{\Tilde{G} \in \mathcal{D}_{support}}-\mathbb{E}\left[\sum_{t}\log p_{MSTM}(\pi t|f(\Tilde{G}))\right]
\end{equation}
where $p_{MSTM}(\pi t|\tilde{G})=softmax(\bm{W}_p\tilde{\bm{h}}_{\pi t}^{(K_2)} + \bm{b}_p)$. Different from previous strategies, which randomly mask atoms or edges to predict the attributes, our method randomly masks some subgraphs and predicts their indices in the vocabulary $\mathbb{V}$ with the proposed decomposition-polymerization architecture. Actually, our prediction task is more difficult since it operates on subgraphs and the size of $\mathbb{V}$ is larger than the size of atom types. As a result, the learned substructure-aware representation captures high-level semantics of substructures and their interactions and can be better generalized to the combinations of known subgraphs under different scaffolds.

\paragraph{Graph-level self-supervised learning}
Node-level  pre-training is not sufficient to acquire generalizable features~\cite{xia2023mole}. Therefore, we propose graph-level self-supervised learning as shown in Eq.\ref{eq8}, where $\mathcal{B}^-$ denotes the negative samples for the anchor sample $G_i$. 
\begin{equation}
    \mathcal{L}_{cl} =\frac{1}{N_2} \sum_{G_i \in \mathcal{D}_{support}} -\log \frac{\exp{(\bm{z}_{S_i}\cdot\bm{z}_{A_i})} }{\exp{(\bm{z}_{S_i}\cdot\bm{z}_{A_i})} + \sum_{G_j \in \mathcal{B}^-}\exp{(\bm{z}_{S_i}\cdot\bm{z}_{A_j})}}
    \label{eq8}
\end{equation}
We denote $\bm{z}_A$ and $\bm{z}_S$ as output features of atom-wise and subgraph-wise branches, respectively. The atom-wise branch and subgraph-wise branch focuses on the different viewpoint of the same molecule. Our method maximizes the feature consistency along these two branches and improves the generalization for downstream tasks. In addition, graph-level self-supervised learning makes the two branches interact which can utilize the superiority of our bilateral architecture.

\newtheorem{Proposition}{Proposition}
\newtheorem{Theorem}{Theorem}
\newtheorem{Corollary}{Corollary}

\subsection{Analysis}
\label{ana}
In this section, we demonstrate that our ASBA has a lower error bound compared with the single atom-wise branch. For simplicity, we consider the case of 1-dimensional feature space, and the conclusion can also be applied in multi-dimensional feature space~\cite{tumer1996linear}. Firstly, we give the upper bound of the Bayes error: 
\begin{Theorem}
\label{the1}
For a binary classification problem $T$, we denote $\mu_0$, $\mu_1$ as the mean vector and $\Sigma_0$, $\Sigma_1$ as the covariance matrix. Let $\Sigma$ be the non-singular, average covariance matrix and $\Sigma=p(c_1)\Sigma_1 + p(c_2)\Sigma_2$, where $P(c_1)$ and $P(c_2)$ are the prior class probability for class $c_1$ and $c_2$, respectively. Then we provide the upper bound of Bayes error $\mathcal{R}_T$~\cite{jain2000statistical}:
\begin{equation}
    \mathcal{R}_T \leq \frac{2P(c_1)P(c_2)}{1+P(c_1)P(c_2)\Delta}, \quad where \quad \Delta=(\mu_1-\mu_2)^T\Sigma^{-1}(\mu_1-\mu_2)
\end{equation}
\end{Theorem}

Based on the analysis in~\cite{tumer1996linear}, we give the upper bound of the expected error for the atom-wise branch and our ASBA in Theorem~\ref{the2} and Theorem~\ref{the3}, respectively. Proofs are presented in the appendix.
\begin{Proposition}
\label{pro2}
Given a single classification model $f$, we have $f(c|x) = p(c|x) + \epsilon_{c, f}(x)$, where $\epsilon_{c, f}(\cdot)$ is the inherent error associated with classes and models. We define $\epsilon_{c, f}(x) = \beta_{c}+\eta_{c, f}(x)$, where $\beta_{c}$ is a constant but $\eta_{c, f}(\cdot)$ is a function associated with samples. 
\end{Proposition}
\begin{Theorem}
\label{the2}
Given a single classification model $f$, the upper bound of expected error $\mathcal{R}_T(f)$ is:
\begin{equation}
    \mathcal{R}_T(f) = \mathcal{R}_T + \frac{\sigma_{\eta_{c_1, f}}^2 + \sigma_{\eta_{c_2, f}}^2}{2|\nabla_{x=x^*} p(c_1|x) - \nabla_{x=x^*} p(c_2|x)|}
\end{equation}
\end{Theorem}

\begin{Theorem}
\label{the3}
Given two classification models $f$, and $g$,  which denote atom-wise branch and subgraph-wise branch respectively, we ensemble the logits as the final output. The upper bound of expected error  $\mathcal{R}_T(f+g)$ is:
\begin{equation}
    \mathcal{R}_T(f+g) = \mathcal{R}_T + \frac{\sigma_{\eta_{c_1, f}}^2 + \sigma_{\eta_{c_2, f}}^2 + \sigma_{\eta_{c_1, g}}^2 + \sigma_{\eta_{c_2, g}}^2}{8|\nabla_{x=x^*} p(c_1|x) - \nabla_{x=x^*} p(c_2|x)|}
\end{equation}
\end{Theorem}

Both $\mathcal{R}_T(f)$ and $\mathcal{R}_T(f+g)$ are closely associated with the inherent error $\eta$. We give a rational hypothesis that inherent errors of two classification models are bounded and the performance of the atom-wise branch is not much worse than that of the subgraph-wise branch. We can draw the conclusion that ASBA has a lower error bound compared with the atom-wise branch. The proof is presented in the appendix.
\begin{Corollary}
\label{coro1}
When satisfying $-\tau_1<\eta_{c, f} < \tau_1$ and $-\tau_2<\eta_{c, g} < \tau_2$, i.e., the error of both atom-wise and subgraph-wise models are bounded. If satisfying $\tau_2 < \sqrt{3}\tau_1$, the expected error upper bound of $\mathcal{R}_T(f+g)$ will be smaller than single atom-wise branch expected error upper bound
we have $\mathcal{R}_T(f)$.
\end{Corollary}

%% file: 3.experiment.tex
\section{Experiments}

\begin{table*}[t]
\caption{Test ROC-AUC performance of different methods on molecular property classification tasks. AVG represents the average results over all benchmarks. We highlight the best and second-best results with $\bm{bold}$ and $\underline{bold}$. We report the mean and standard results. }
\scriptsize
\center
\begin{tabular}{lcccccccc|r}
\toprule
Methods & BACE &BBBP & ClinTox & HIV & MUV & SIDER & Tox21 & ToxCast &Avg \\
\midrule
Infomax& 75.9(1.6)& 68.8(0.8)& 69.9(3.0)& 76.0(0.7)& 75.3(2.5)& 58.4(0.8)& 75.3(0.5)& 62.7(0.4)& 70.3\\
AttrMasking& 79.3(1.6)& 64.3(2.8)& 71.8(4.1)& 77.2(1.1)& 74.7(1.4)& 61.0(0.7)& \underline{76.7(0.4)}& 64.2(0.5)& 71.1\\
GraphCL& 75.4(1.4)& 69.7(0.7)& 76.0(2.7)& 78.5(1.2)& 69.8(2.7)& 60.5(0.9)& 73.9(0.7)& 62.4(0.6)& 70.8\\
AD-GCL& 78.5(0.8)& 70.0(1.1)& 79.8(3.5)& 78.3(1.0)& 72.3(1.6)& 63.3(0.8)& 76.5(0.8)& 63.1(0.7)& 72.7\\
MGSSL& 79.1(0.9)& 69.7(0.9)& \underline{80.7(2.1)}& \underline{78.8(1.2)}& \underline{78.7(1.5)}& 61.8(0.8)& 76.5(0.3)& 64.1(0.7)& 73.7\\
GraphLoG& \textbf{83.5(1.2)}& \underline{72.5(0.8)}& 76.7(3.3)& 77.8(0.8)& 76.0(1.1)& 61.2(1.1)& 75.7(0.5)& 63.5(0.7)& 73.4\\
GraphMVP& 81.2(0.9)& 72.4(1.6)& 77.5(4.2)& 77.0(1.2)& 75.0(1.0)&\textbf{63.9(1.2)}& 74.4(0.2)& 63.1(0.4)& 73.1\\
GraphMAE& \underline{83.1(0.9)}& 72.0(0.6)& \textbf{82.3(1.2)}& 77.2(1.0)& 76.3(2.4)& 60.3(1.1)& 75.5(0.6)& 64.1(0.3)& \underline{73.8}\\
\midrule
ASBA(node) & 81.9(0.2) &71.8(3.0) &79.0(2.2) &\textbf{78.8(0.6)} &73.7(3.4) &61.2(0.5) &\textbf{76.7(0.2)} &\underline{65.6(0.4)} & 73.6 \\
ASBA(node+graph) & 79.4(2.1) &\textbf{73.9(0.4)} &79.5(1.6) &76.3(0.4) &\textbf{78.7(1.2)} &\underline{63.4(0.4)} &75.8(0.3) &\textbf{65.7(0.1)} & \textbf{74.1} \\
\bottomrule
\end{tabular}
\label{tab1}
\end{table*}

\begin{table*}[t]
\caption{Test RMSE performance of different methods on the regression datasets.}
\center
\scriptsize
\begin{tabular}{lccc|ccc}
\midrule
\multirow{3}{*}{Methods} & \multicolumn{6}{c}{Regression dataset}\\ 
\cmidrule{2-7}
& \multicolumn{3}{c|}{fine-tuning} & 
\multicolumn{3}{c}{linear probing}\\
 & FreeSolv & ESOL & Lipo  & FreeSolv & ESOL & Lipo \\
\midrule
\multicolumn{1}{l}{Infomax} & 3.416(0.928)& 1.096(0.116)& 0.799(0.047) & 4.119(0.974) & 1.462(0.076) & 0.978(0.076)  \\
\multicolumn{1}{l}{EdgePred} &3.076(0.585)& 1.228(0.073)& 0.719(0.013) &3.849(0.950) & 2.272(0.213) & 1.030(0.024)  \\
\multicolumn{1}{l}{Masking} &3.040(0.334)& 1.326(0.115)& 0.724(0.012) &3.646(0.947) & 2.100(0.040) &1.063(0.028) \\
\multicolumn{1}{l}{ContextPred} &2.890(1.077)& 1.077(0.029)&  0.722(0.034) & 3.141(0.905) & \underline{1.349(0.069)} & 0.969(0.076)  \\
\multicolumn{1}{l}{GraphLog} &2.961(0.847)& 1.249(0.010)& 0.780(0.020) & 4.174(1.077) & 2.335(0.073) & 1.104(0.024)  \\
\multicolumn{1}{l}{GraphCL} &3.149(0.273)& 1.540(0.086)&  0.777(0.034) & 4.014(1.361) & 1.835(0.111) & 0.945(0.024)  \\
\multicolumn{1}{l}{3DInfomax} &\underline{2.639(0.772)}& \textbf{0.891(0.131)}& 0.671(0.033) & \underline{2.919(0.243)} & 1.906(0.246) & 1.045(0.040)  \\
\multicolumn{1}{l}{GraphMVP} &2.874(0.756)& 1.355(0.038)& 0.712(0.025)& \textbf{2.532(0.247)} & 1.937(0.147) & 0.990(0.024)  \\
\midrule
\multicolumn{1}{l}{ASBA(node)} &\textbf{2.589(0.665)}  &1.050(0.116) &\underline{0.562(0.021)}   &3.237(0.707)  &1.397(0.136) &\underline{0.807(0.041)}  \\
\multicolumn{1}{l}{ASBA(node+graph)} &3.089(0.841) &\underline{1.042(0.106)} &\textbf{0.546(0.001)} &3.318(0.708) &\textbf{1.296(0.350)} &\textbf{0.692(0.011)}  \\

\bottomrule
\end{tabular}
\label{tab2}
\vskip -0.1in
\end{table*}

\subsection{Datasets and experimental setup}
\paragraph{Datasets and Dataset Splittings} 
We use the ZINC250K dataset \cite{sterling2015zinc} for self-supervised pre-training, which is constituted of $250k$ molecules up to $38$ atoms. As for downstream molecular property prediction tasks, we test our methods on $8$ classification tasks and $3$ regression tasks from MoleculeNet~\cite{wu2018moleculenet}. For classification tasks, we follow the \emph{scaffold-splitting}~\cite{ramsundar2019deep}, where molecules are split according to their scaffolds (molecular substructures). The proportion of the number of molecules in the training, validation, and test sets is $80\%:10\%:10\%$. Following~\cite{li2022kpgt}, we apply random scaffold splitting to regression tasks, where the proportion of the number of molecules in the training, validation, and test sets is also $80\%:10\%:10\%$. Following~\cite{zhang2021motif, liu2021pre}, we performed three replicates on each dataset to obtain the mean and standard deviation.

\paragraph{Model configuration and implemented details}
To verify the effectiveness of our ASBA, we do experiments with different molecular fragmentation methods, such as BRICS~\cite{degen2008art} and the principle subgraph~\cite{kong2022molecule}. We also do experiments with different hyper-parameter $K_1$ and $K_2$. For self-supervised learning, we employ experiments with the principle subgraph with vocabulary size $|\mathbb{V}|=100$. We perform the subgraph embedding module and the subgraph-wise polymerization module with a $K_1=2$ layer GIN~\cite{leskovec2019powerful} and a $K_2=3$ layer GIN, respectively. For downstream classification tasks and regression tasks,  we mainly follow previous work~\cite{hu2019strategies} and~\cite{li2022kpgt}, respectively. For the pre-training phase, we follow the work~\cite{zhang2021motif} to use the Adam optimizer~\cite{kingma2014adam} with a learning rate of $1 \times 10^{-3}$, and batch size is set to $32$.

\paragraph{Baselines}
For classification tasks, we comprehensively evaluated our method against different self-supervised learning methods on molecular graphs, including Infomax~\cite{velivckovic2018deep}, AttrMasking~\cite{hu2019strategies}, ContextPred \cite{hu2019strategies}, GraphCL~\cite{you2020graph}, AD-GCL~\cite{suresh2021adversarial}, MGSSL~\cite{zhang2021motif}, GraphLog~\cite{xu2021self}, GraphMVP~\cite{liu2021pre} and GraphMAE~\cite{hou2022graphmae}. For regression tasks, we compare our method with Infomax~\cite{velivckovic2018deep}, EdgePred~\cite{hamilton2017inductive}, AttrMasking~\cite{hu2019strategies}, ContextPred \cite{hu2019strategies}, GraphLog~\cite{xu2021self}, GraphCL~\cite{you2020graph},3DInformax~\cite{stark20223d} and GraphMVP~\cite{liu2021pre}. Among them, 3DInfomax exploits the three-dimensional structure information of molecules, while other methods also do not use knowledge or information other than molecular graphs.


\begin{table*}[t]
\caption{Test ROC-AUC performance of different methods on molecular property classification tasks with different tokenization algorithms and model configurations. }
\scriptsize
\center
\begin{tabular}{lcccccccc|r}
\toprule
Methods & BACE &BBBP & ClinTox & HIV & MUV & SIDER & Tox21 & ToxCast &Avg \\
\midrule
Atom-wise&71.6(4.5) &68.7(2.5) &57.5(3.8) &75.6(1.4) &73.2(2.5) &57.4(1.1) &74.1(1.4)  &62.4(1.0)& 67.6\\
\midrule
\multicolumn{9}{c}{The principle subgraph, $|\mathbb{V}|=100$, $K_1=2$, $K_2=3$}\\
Subgraph-wise& 64.4(5.2)& 69.4(3.0)& 59.2(5.2)&71.7(1.5) & 68.3(1.6) &59.1(0.9) & 72.6(0.8) &61.3(0.7)& 65.8\\
ASBA& 72.2(3.3)& 70.6(3.2)& 61.1(3.0)&76.1(1.8) & 72.8(3.3) &59.8(1.3)& 75.1(0.4) &63.7(0.5)& 69.0\\
\midrule
\multicolumn{9}{c}{The principle subgraph, $|\mathbb{V}|=100$, $K_1=3$, $K_2=2$}\\
Subgraph-wise& 63.1(7.1) &68.7(2.7) &55.8(6.3) &71.7(1.7) &68.3(4.9) &58.7(1.5) &72.9(0.9)  &61.5(0.9)& 65.1\\
ASBA& 72.3(2.1) &70.9(2.5) &60.6(4.9) &75.5(1.3) & 73.3(2.2) &59.6(0.7) &75.3(0.7) &63.9(0.5) &68.9 \\
\midrule
\multicolumn{9}{c}{The principle subgraph, $|\mathbb{V}|=300$, $K_1=2$, $K_2=3$}\\
Subgraph-wise& 66.2(4.5)& 63.7(3.2)& 59.0(8.5)&74.2(1.6) &68.9(1.9) &61.6(1.8) &73.3(0.9) &60.5(0.5) &65.9 \\
ASBA& 71.2(4.1)& 66.0(4.4)& 59.7(4.7)&77.7(1.7) &72.7(2.9) &61.4(1.0) &75.9(0.4) &63.6(0.8) &68.5\\
\midrule
\multicolumn{9}{c}{The principle subgraph, $|\mathbb{V}|=300$, $K_1=3$, $K_2=2$}\\
Subgraph-wise&67.3(2.0) &66.5(3.3) &54.7(5.7) &73.8(2.1) &69.7(2.5) &60.8(2.5) &73.7(0.7)  &60.8(0.7)& 65.9\\
ASBA &71.8(2.5)&68.9(2.3) &55.9(5.3) &77.4(1.8) &74.7(2.5) &61.0(1.1) &76.4(0.7) &63.7(0.6)& 68.7\\
\midrule
\multicolumn{9}{c}{BRICS, $K_1=2$, $K_2=3$}\\
Subgraph-wise&71.4(3.9) &66.1(3.5) &51.8(3.7) &75.2(1.8) &70.0(2.0) &56.0(1.5) &74.0(0.8)  &64.2(1.1)& 66.1\\
ASBA&74.3(4.5) &69.3(2.8) &55.9(3.1) &76.7(1.6) &75.0(2.1) &58.5(1.5) &76.2(0.5) &64.9(1.2)& 68.9 \\
\midrule
\multicolumn{9}{c}{BRICS, $K_1=3$, $K_2=2$}\\
Subgraph-wise&73.6(3.7) &67.0(1.9) &53.0(5.3) &74.0(1.7) &70.5(1.9) &55.8(1.7) &74.4(1.0)  &65.0(0.4) &66.7\\
ASBA&76.8(3.9) &67.8(3.6) &60.0(3.3) &76.9(1.2) &74.4(2.4) &58.1(1.1) &76.7(0.6) &65.7(0.6) &69.6 \\
\bottomrule
\end{tabular}
\label{tab3}
\end{table*}

\begin{table*}[t]
\caption{Ablation study of self-supervised learning. We report the performance of two branches with different self-supervised learning methods.}
\scriptsize
\center
\begin{tabular}{lcccccccc|r}
\toprule
Methods & BACE &BBBP & ClinTox & HIV & MUV & SIDER & Tox21 & ToxCast &Avg \\
\midrule

\multicolumn{9}{c}{Node-level self-supervised learning}\\
Atom-wise& 80.0(0.8) &66.4(2.0) &76.8(5.6) &78.6(0.4) &72.4(0.8) &58.2(0.9) &75.4(0.8) &64.5(0.4) & 71.5\\
Subgraph-wise& 74.3(2.8) &72.4(1.1) &63.0(7.1) &74.7(1.4) &70.8(4.1) &62.5(0.8) &75.0(0.2) &63.6(0.6) & 69.5 \\
ASBA &81.9(0.2) &71.8(3.0) &79.0(2.2) &78.8(0.6) &73.7(3.4) &61.2(0.5) &76.7(0.2) &65.6(0.4) & 73.6 \\

\midrule
\multicolumn{9}{c}{Node-level + Graph-level self-supervised learning}\\
Atom-wise& 76.4(2.5) &68.1(1.9) &79.3(1.4) &76.4(0.4) &76.4(0.3) &59.3(1.1) &74.2(0.4) &64.2(0.2) & 71.8\\
Subgraph-wise& 77.8(0.9) &71.9(3.4) &75.8(2.9) &73.4(0.3) &74.8(1.2) &63.8(0.8) &74.5(0.4) &63.3(0.7) &71.9  \\
ASBA &79.4(2.1) &73.9(0.4) &79.5(1.6) &76.3(0.4) &78.7(1.2) &63.4(0.4) &75.8(0.3) &65.7(0.1) & 74.1 \\





\bottomrule
\end{tabular}
\label{tab4}
\end{table*}

\subsection{Results and Analysis}
\paragraph{Classification}
Tab.~\ref{tab1} presents the results of fine-tuning compared with the baselines on classification tasks. ``ASBA(node)'' denotes we apply node-level self-supervised learning method on our ASBA while  ``ASBA(node+graph)'' denotes we combine node-level and graph-level self-supervised learning methods. 
From the results, we observe that the overall performance of our ``ASBA(node+graph)'' method is significantly better than all baseline methods, including our ``ASBA(node)'' method, on most datasets. Among them, AttrMasking and GraphMAE also use masking strategies which operate on atoms and bonds in molecular graphs. Compared with AttrMasking, our node+graph method achieves a significant performance improvement of 9.6\%, 7.7\%, and 4\% on BBBP, ClinTox, and MUV datasets respectively, with an average improvement of 3\% on all datasets. Compared with GraphMAE, our method also achieved a universal improvement. Compared with contrastive learning models, our method achieves a significant improvement with an average improvement of 3.8\% compared with Infomax, 3.3\% compared with GraphCL, 1.4\% compared with AD-GCL, and 0.7\% compared with GraphLoG. For GraphMVP which combines contrastive and generative methods, our method also has an average improvement of 1\%. 

\paragraph{Regression}
In Tab.~\ref{tab2}, we report evaluation results in regression tasks under the fine-tuning and linear probing protocols for molecular property prediction. Other methods are pre-trained on the large-scale dataset ChEMBL29~\cite{gaulton2012chembl} containing 2 million molecules, which is 10 times the size of the dataset for pre-training our method. The comparison results show that our method outperforms other methods and achieves the best or second-best performance in five out of six tasks, despite being pre-trained only on a small-scale dataset. This indicates that our method can better learn transferable information about atoms and subgraphs from fewer molecules with higher data-utilization efficiency.

\paragraph{ASBA can achieve better generalization}
In Tab.~\ref{tab3}, we compare the performance of our atom-level, subgraph-level, and our ASBA integrated model on different classification tasks. From the experimental results, it can be seen that the atom-level branch performs better than the subgraph-level branch on some datasets, such as BACE, ToxCast, and Tox21, while the subgraph-level branch outperforms on others, such as SIDER and BBBP. This is because the influencing factors of different classification tasks are different, some focus on functional groups, while some focus on the interactions between atoms and chemical bonds. However, no matter how the parameters of the model change, our ASBA always achieves better results than the two separate branches on all datasets since it integrates the strengths of both. These results demonstrate that our ASBA has better generalization. We also visualize the testing curves in Fig.~\ref{fig1} (b) and (c), the aggregation results outperform both atom-wise and subgraph-wise  in most epochs. 

\paragraph{The effectiveness of graph-level self-supervised learning in the pre-training stage }
From Tab.~\ref{tab1} and Tab.~\ref{tab2}, it can be seen that adding graph-level self-supervised learning
in the pre-training stage can achieve better results in most tasks. Therefore, we investigate two different pre-training methods in Tab.~\ref{tab4}. From the experimental results, it can be seen that pre-training the two branches jointly utilizes two self-supervised learning methods than pre-training the two branches individually (only applying node-level self-supervised learning) in general. For detail, graph-level self-supervised learning can give 0.3\% and 2.4\% gains for the atom-wise branch and subgraph-wise branch, respectively. The overall performance improved by 0.5\%. These results demonstrate the necessity of applying graph-level self-supervised learning, which realizes the interaction between two branches.


%% file: 4.conclusion.tex
\section{Conclusion}
In this paper, we address the limitation that molecular properties are not solely determined by atoms by proposing a novel approach called Atomic and Subgraph-aware Bilateral Aggregation (ASBA). The ASBA model consists of two branches: one for modeling atom-wise information and the other for subgraph-wise information. To improve generalization, we propose a node-level self-supervised learning method called MSTM for the under-explored subgraph-wise branch. Additionally, we introduce a graph-level self-supervised learning method to facilitate interaction between the atom-wise and subgraph-wise branches. Experimental results show the effectiveness of our method. 

\textbf{Limitation:} It is important to note that we did not conduct self-supervised learning experiments on a larger unlabeled dataset due to resource and time limitations.